\documentclass[10pt,twocolumn,letterpaper]{article}

\usepackage{cvpr}              

\usepackage{graphicx}
\usepackage{amsmath}
\usepackage{amssymb}
\usepackage{booktabs}

\usepackage{multirow}
\usepackage{mathrsfs}
\usepackage{enumitem}
\usepackage{makecell}
\usepackage{tabulary}
\usepackage{pifont}
\usepackage{bm}

\usepackage[pagebackref,breaklinks,colorlinks]{hyperref}

\usepackage[capitalize]{cleveref}
\crefname{section}{Sec.}{Secs.}
\Crefname{section}{Section}{Sections}
\Crefname{table}{Table}{Tables}
\crefname{table}{Tab.}{Tabs.}

\def\cvprPaperID{3408} 
\def\confName{CVPR}
\def\confYear{2022}

\begin{document}

\title{AutoCaption: Image Captioning with Neural Architecture Search}

\author{Xinxin Zhu$^1$, Weining Wang$^1$, Longteng Guo$^{1,2}$ and Jing Liu$^{1,2}$\\
{$^1$ National Lab of Pattern Recognition, Institute of Automation, Chinese Academy of Sciences}\\
{$^2$ School of Artificial Intelligence, University of Chinese Academy of Sciences}\\
{\tt\small \{xinxin.zhu, weining.wang, longteng.guo, jliu\}@nlpr.ia.ac.cn}
}
\maketitle

\begin{abstract}
   Image captioning transforms complex visual information into abstract natural language for representation, which can help computers understanding the world quickly. However, due to the complexity of the real environment, it needs to identify key objects and realize their connections, and further generate natural language. The whole process involves an encoder module for visual understanding and a decoder module for language generation, which brings more challenges to the design of deep neural networks than other tasks. In this paper, we adopt a strategy of Network Architecture Search (NAS) to better design the RNN-based decoder module for image captioning automatically, called as AutoCaption. Specially, the reinforcement learning method based on shared parameters are explored to design the AutoCaption Model efficiently, and the search space includes connections between the layers and the operations in layers both in order to generate more architectures for the decoder. It is noted that RNN is equivalent to a subset of our search space. Experiments on the MSCOCO datasets show that our Auto-Caption model can achieve better performance than traditional hand-design methods.

\end{abstract}

\section{Introduction}

Image captioning is an important technology to let the computer understand the semantic meaning of the image and describe the image with the natural language further.
It is a multi-modal problem using computer vision and natural language processing technologies.
Nowadays, a variety of computer vision algorithms have emerged to help computers understand the real world, such as image classification \cite{he2016deep,Krizhevsky2012ImageNet}, object detection \cite{ren2015faster,Redmon2016You,Liu2015SSD}, image segmentation \cite{long2015fully,U-Net,Chen2016DeepLab} and other technologies. 
Through these algorithms, computers can detect and understand objects in the real world, but they cannot express the relations between objects.
Image captioning can understand of key objects and their attributes in visual information. 
Then, it should to further understand the connections between objects. 
At last, it can describe the image with fluent natural language.
In order to realize image understanding and text generation, the model includes the image understanding module and the language generation module.
Traditional deep learning-based methods are all manually designed by relevant experts based on experience, and the process is tedious and complicated.
The structure of the deep neural network has an important impact on the final performance, but the discovery of the new neural network structure requires much effort from experts, and the parameters need to be adjusted continuously for the data, which brings specific challenges to the development of image captioning.

Network Architecture Search (NAS) has achieved great success in image recognition. 
It helps us discover better models in computer vision tasks, such as
image classification \cite{cai2018proxylessnas}, 
semantic segmentation \cite{Liu_2019_CVPR} and object detection \cite{Ghiasi_2019_CVPR}. 
However, the research on image captioning is still slightly insufficient.
Some efforts have been invested in searching for sequence models.
\cite{Zoph17a,enas} use NAS to find better RNN architectures for text translation, 
and \cite{so2019evolved} employs evolution algorithms to search better transformer architectures.
\cite{Zoph17a} requires a lot of computing resources, which is difficult for most researchers. 
In this work, we propose a more efficient yet effective methodology to improve the text generation of the image captioning. 

With the development of neural architecture search methods in the past years, a series of automatically designed image recognition and natural language processing-related network structures have emerged, which have exceeded the networks designed by humans in some respects.
The neural network automatic design of image captioning is more challenging than the image recognition task. 
The algorithm is composed of two different neural networks, a vision understanding module and a language generation module, which brings challenges to the automatic design of neural networks. 
This paper is oriented to the task of image captioning. 
Through the network architecture search technology, it automatically designs the language generation module for image captioning, and can dynamically adjust the computational complexity of the neural network according to the specific task.
Transformer shows excellent performance on image captioning tasks \cite{xlinear}.
It was originally invented for machine translation and other natural language tasks, and various improved models, including GPT, appeared. 
These models are more complex than LSTM and can handle more complex language words relationships. 
However, image captioning is an effective way to solve these problems. 
The sentences used in the task are relatively short, usually less than 20 words. 
If the more complex transformer model is used directly, it will increase unnecessary calculation cost.

In this work, we experiment on making the design choices in the image captioning generation model automatically, i.e., whether to scale, where to connection, number of blocks, activation function, etc., so that we can obtain a better text generation architecture that better suits the image caption task. 
In order to better search our structure, we employ reinforcement learning (RL)
and we use the specialized parameter sharing strategies for AutoRNN blocks to help speed up the search process. 
Furthermore, during the training process, we use the evaluation metric to sample the best architecture with the controller which has been trained.

Experiments on the MSCOCO dataset shows that the automatically designed image captioning models can outperform the standard transformer models significantly.

To summarize, the main contributions of this study is three-fold:
\begin{itemize}
\item [(1)] We put forward a neural architecture search (NAS) framework for image captioning tasks to find better text generation networks.
Compared with other image captioning models based on the transformer or LSTM, the proposed model can achieve better performance.

\item [(2)] We use the reinforcement learning method based on shared parameters for automatic design of text generation method of image captioning efficiently.
We first use the relatively simple show tell \cite{vinyals2017show} model to search the network architecture.
Then we integrate the search model with the bilinear attention \cite{xlinear} to further verify the performance of the model.

\item [(3)] We conduct extensive experiments on the MSCOCO benchmark datasets.
The proposed Auto-Caption model achieves better performance than X-Linear \cite{xlinear} benchmark method.
\end{itemize}

\section{Related Work}

\subsection{Image Captioning}
With the development of deep learning, image captioning have developed rapidly.
Machine translation tasks \cite{Cho2014Learning} and sequence-to-sequence tasks \cite{Sutskever:NIPS14} have also begun to use neural network methods, and their effects are much better than traditional methods, which brings new inspiration to the language generation module in image captioning. 
In 2014, Kiros et al. \cite{kiros2014unifying} firstly proposed a method of image captioning method based on deep neural network, and merged the image and its description text with other multi-modal information to generate a description sentence successfully. 
Mao et al. \cite{mao2014deep} proposed a multi-modal recurrent neural network (m-RNN). 
The network includes two subnets: deep convolution neural network for image feature extraction and deep recurrent neural network for text encoding.
Vinyals et al. \cite{Vinyals14} proposed an encoder-decoder framework, which uses Convolution Neural Network (CNN) as the encoder and Long Short-Term Memory (LSTM) \cite{Hochreiter:NC97} as the decoder which is used to generate sentences describing the image. 
Xu et al. \cite{xu2015show} proposed an image captioning model based on attention mechanism. 
Their model includes three parts, encoder, decoder and attention model. 
The encoder is used to encode the visual information into visual feature, and the decoder is used to decode the visual information into descriptive text, and the attention mechanism is used to guide the model to focus on the image content generation stage of a specific area during the text. 
These methods all use the RNN as the decoder to generate descriptions.
However, the above decoders of the image captioning model are all artificially designed networks, not necessarily the best choice for image captioning tasks.

\begin{figure*}[!tb]
\centering
\includegraphics[width=5in,height=2.8in]{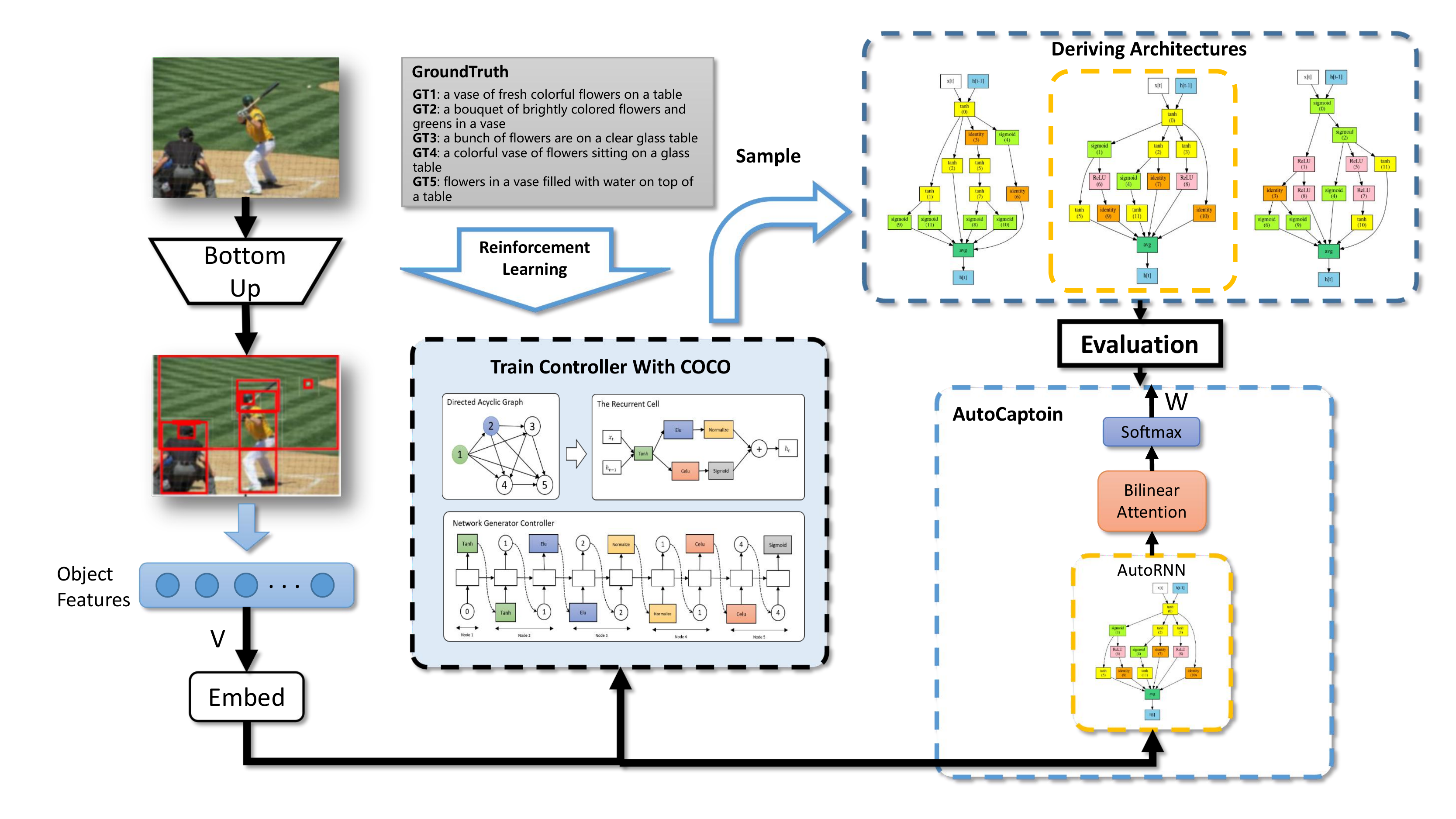}
\caption{
Overview of our AutoCaption for image captioning. 
Faster R-CNN is firstly utilized to detect a set of image regions. 
Next, AutoRNN is automatically designed by the reinforcement learning method. 
Then, we sample several models by the controller which is been trained in the MSCOCO dataset.
We choose the network which performs best in the CIDEr metric as our the final AutoRNN network.
At last, we integrate the AutoRNN with bilinear attention \cite{xlinear} to evaluate the proposed model.}
\label{fig:framework}
\end{figure*}

\subsection{Neural Architecture Search}
In 2016, MIT and Google proposed the network architecture search \cite{Baker17,Zoph17a} to design a deep neural network structure automatically respectively. 
These models defeated the models designed by humans and the possibility of automatic design model was proved.
In 2017, Google proposed a network structure search algorithm based on evolutionary algorithms \cite{Real17}. 
The evolutionary algorithm is based on Darwin's biological evolution theory, which is more in line with the laws of natural selection and survival of the fittest in nature. 
The network architectures designed by the automatic method have made breakthroughs in image detection and image segmentation \cite{autofpn,detnas,peng2019enat,auto_deeplab}.
It can be seen from the research in recent years that NAS has become a research hotspot and has shown advantages over traditional hand-designed networks in many fields.
Image captioning also urgently need an automatic network design method to design a more effective network for image understanding and text generation in image captioning.
A few previous studies have attempted to optimize existing conventional recurrent cells and concluded that they are not necessarily optimal, moreover, the significance of their components is unclear\cite{greff2016lstm,jozefowicz2015empirical}.
Greff et al. \cite{greff2016lstm} have make some modifications of LSTM (Long-Short Term Memory \cite{hochreiter1997long}) on several tasks. 
Jozefowich et al. \cite{jozefowicz2015empirical} tried to find an architecture that outperforms LSTM using evolutionary approach applied to GRU (Gated Recurrent Unit \cite{cho2014properties}) and LSTM.
There are also the works that introduce algorithms to search RNN cells \cite{zoph2016neural,liu2018darts,pham2018efficient}. 
However, these above works are all pure text tasks, and have not been effectively designed in a multi-modal situation.

\section{AutoCaption}

The overall of our model is presented in Figure \ref{fig:framework}.
Faster R-CNN is firstly utilized to detect a set of image regions. 
Then, AutoRNN is automatically designed by the reinforcement learning method. 
After that, we sample several models by the controller which is been trained in the MSCOCO dataset.
We choose the network which performs best in the CIDEr metric as our final AutoRNN network.
At last, we integrate the AutoRNN with bilinear attention \cite{xlinear} to evaluate the proposed model.

Existing text generation models are usually based on the LSTM model or Transformer model, and these two important models were originally used for natural language processing tasks such as machine translation.
Image captioning needs to generate sentences that can describe key contents according to key objects and their attributes, so different requirements are put forward for the generation of natural language.
The sentence in the image captioning is mainly based on the objective statement semantics, which has the necessary modifications to the main object and the secondary object, and contains the state or motion information of the main object, so the method of directly applying the traditional model contains too many redundant parameters also affect the operating efficiency of the model.
Network architecture search (NAS) can describe the characteristics of data distribution according to image captioning, and design text generation module more effectively.
Network architecture search includes two important parts, search space and search strategy. For the text generation task itself, the search space refers to the traditional RNN model, which mainly includes the basic search unit such as the connection, the nonlinear activation layer and the control gate.

In order to improve the efficiency of the architecture search, we use the classic showtell \cite{vinyals2017show} as our search framework.
Compared with other models, this method only retains the basic encoder and decoder.
We use bottom-up \cite{anderson2017bottom} as our encoder.
This method uses the object detection method pre-trained on Visual Genome \cite{krishna2017visual} as the image feature extractor.
Since the bottom-up features are extracted in advance, this solves a lot of search time.
The entire search is mainly focused on the search for the decoder part.
The decoder is a text generation task and it is a sequence task. 
Some works \cite{huang2019attentio,xlinear} have shown that image captioning model with the transformer has shown better performance than the model with the LSTM, 
but this will further bring a lot of computational overhead.

\begin{figure}[!tb]
\centering
\includegraphics[width=3in]{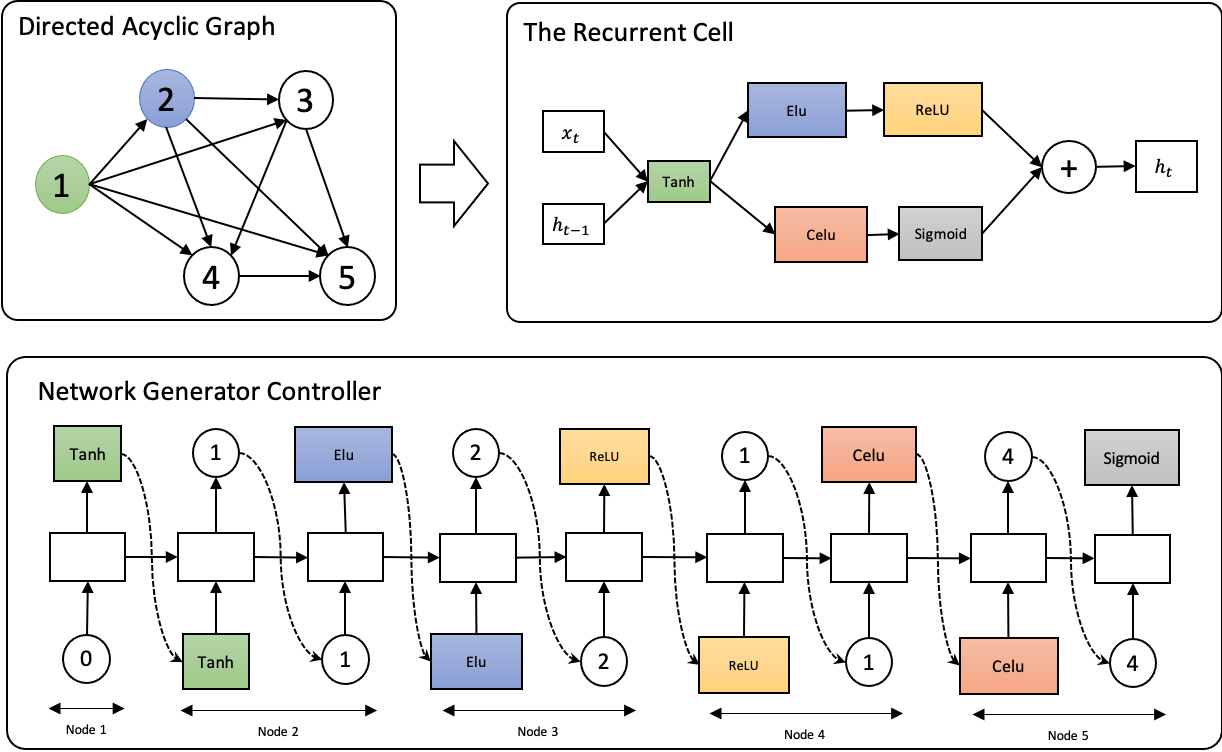}
\caption{Overview of the AutoRNN. We employ a DAG with $N$ nodes to design the recurrent cells,
where the modes represent computations of the layer and
the edges represent the connection between the nodes.
We use the LSTM as our the controller to design the the recurrent cells for image captioning model.}
\label{fig:autornn}
\end{figure}

\subsection{Search Space Design}

We employ a DAG with $N$ nodes to design the recurrent cells,
where the modes represent computations of the layer and
the edges represent the connection between the nodes.
We use the LSTM as our the controller to design the the recurrent cells for image captioning model.
We call the searched recurrent cells as AutoRNN.
Different from the RNN cells in \cite{zoph2016neural},
where the authors fix the topology of their architectures as a binary tree and only learn the operations at each node of the tree.
In contrast, our search space allows AutoRNN to design both the topology and the operations in RNN cells, and hence is more flexible.

Now we will discuss the search space in detail.
Our goal here is to optimize the text generation module of the image captioning model.
We used the RNN architecture as our text generation module and our goal here is to optimize the RNN architecture.
We maintain the basic structure of RNN. 
The search layers and search space are list here.

\begin{itemize}
    \item Activation = {relu, tanh, sigmoid, elu, celu, gelu, leaky\_relu, silu};
    \item Number of layers  = {6, 8, 10, 12};
	\item The model embedding size = {200, 512, 1000, 2048};
	\item The hidden state size of the model = {200, 512, 1000, 2048};
	\item The label smoothing value = {0, 0.1};
	\item Init hidden state every epoch = {True, False};
	\item Use the shared embedding weights on both the input and the output of the model = {True, False};
	\item The hidden state of the controller = {100, 200, 512, 1024};
\end{itemize}

The activation functions we consider are listed in Table \ref{tab:activation}.
We use a variety of activation functions including $Tanh$, $Sigmoid$ used by LSTM, which can expand our search space to find better architectures.
Next, we will show how to obtain a better architecture than standard RNN with this huge search space.

\begin{table}
\centering
\resizebox{0.48\textwidth}{!}{
\begin{tabular}{lcc}
\hline \bf Name  & \bf Function  \\ 
\hline
relu  &  $\max(x, 0)$    \\
tanh & $\frac{\exp(x) - \exp(-x)} {\exp(x) + \exp(-x)}$ \\
sigmoid & $\frac{1}{1 + \exp(-x)} $ \\
elu   &  $x$ if $x \geq 0$ else $e^{x} - 1$  \\
celu   &  $ max(0, x) + min(0,\alpha*(exp(x/\alpha)-1))$  \\
gelu   & $0.5*x*(1 + erf(x / \sqrt{2}))$  \\
leaky\_relu  &  $x$ if $x \geq 0$ else 1e-2 * $x$    \\ 
silu   & $x * sigmoid(x)$   \\
\hline
\end{tabular}}
\caption{\label{tab:activation}
Activation functions in the search space.}
\end{table}

\subsection{Architecture Search}

We use reinforcement learning methods to search the network structure of image captioning models. 
Figure \ref{fig:autornn} shows the flow of network structure search based on reinforcement learning image captioning text generation model. 
It regards the entire network as a directed acyclic graph, in which nodes represent different layers in the network, such as ReLU, Sigmoid and other layers, that is, the search space of the network.
The larger search space can cover more networks, but at the same time it will make the search difficulty increase exponentially.
The edges of a directed acyclic graph represent the connection between different layers. 
In the original AlexNet and VGG networks, more layers were continuously stacked to achieve better performance, and the connection method was a single connection between adjacent layers. 
Later, with the emergence of models such as ResNet \cite{he2016deep} and DenseNet \cite{2017Densely}, the effectiveness of connection methods such as jump connections and dense connections was verified. 
We use the controller to select the connection so that the architecture can express more connection methods.

We employ a controller to do a guided exploitation in search space, which is similar to NASNet \cite{zoph2018learning} and ENAS~\cite{enas}. 
The controller is an LSTM network with parameters $\theta$. 
The output hidden state is fed into a classifier to decide the action at each step. 
The shared parameters of the child models are denoted by $\omega$, which will be discussed in detail in the next.

The architecture search procedure consists of two interleaving phases.
The first phase trains $\omega$, the shared parameters of the child models, on a pass through the training data set. 
The second phase trains $\theta$, the parameters of the controller, via optimizing the expected reward function using the REINFORCE algorithm~\cite{williams1992simple}.

In order to effectively select the nodes and connection methods in the network, we uses the method of reinforcement learning. 
The Network Generator Controller in Figure \ref{fig:autornn} is a network generation controller that is used to select the nodes and connection methods of the network. 
The controller uses the LSTM model, which has shown excellent performance in sequence tasks and is widely used in natural language processing and speech recognition research.
We regard the generation of the network as a sequence task.
The controller selects the most likely node based on the probability distribution.

To create a recurrent cell, the controller RNN samples $N$ blocks of decisions. 
Here we illustrate the mechanism via a simple example recurrent cell with $N = 4$ computational nodes (visualized in Figure~\ref{fig:autornn}). 
Let $x_t$ be the input signal for a recurrent cell (word embedding), and $h_{t-1}$ be the output from the previous time step.

\begin{table*}[t]\scriptsize
    \centering
    \vspace{-0.1in}
    \caption{\small Performance comparisons on MSCOCO Karpathy test split, where B@$N$, M, R, C and S are short for BLEU@$N$, METEOR, ROUGE-L, CIDEr and SPICE scores. All values are reported as percentage (\%). $^{\sum}$ indicates model ensemble/fusion.}
    \vspace{-0.0in}
    \begin{tabular}{l | c c c c c c c c | c c c c c c c c}
        \Xhline{2\arrayrulewidth}
		  & \multicolumn{8}{c|}{\textbf{Cross-Entropy Loss}} & \multicolumn{8}{c}{\textbf{CIDEr Score Optimization}} \\
		                             & B@1  & B@2  & B@3  & B@4  & M    & R    & C     & S    & B@1  & B@2  & B@3  & B@4  & M    & R    & C     & S  \\	
			\hline \hline
LSTM \cite{Vinyals14}            &   -  &   -  &   -  & 29.6 & 25.2 & 52.6 & 94.0  &  -   &  -   &  -   &  -   & 31.9 & 25.5 & 54.3 & 106.3 &  -   \\
SCST \cite{rennie2017self}       &   -  &   -  &   -  & 30.0 & 25.9 & 53.4 & 99.4  &  -   &  -   &  -   &  -   & 34.2 & 26.7 & 55.7 & 114.0 &  -   \\
LSTM-A \cite{yao2017boosting}    & 75.4 &   -  &   -  & 35.2 & 26.9 & 55.8 & 108.8 & 20.0 & 78.6 &  -   &  -   & 35.5 & 27.3 & 56.8 & 118.3 & 20.8 \\
RFNet \cite{jiang2018recurrent}  & 76.4 & 60.4 & 46.6 & 35.8 & 27.4 & 56.5 & 112.5 & 20.5 & 79.1 & 63.1 & 48.4 & 36.5 & 27.7 & 57.3 & 121.9 & 21.2 \\
Up-Down \cite{anderson2017bottom}& 77.2 &   -  &   -  & 36.2 & 27.0 & 56.4 & 113.5 & 20.3 & 79.8 &  -   &  -   & 36.3 & 27.7 & 56.9 & 120.1 & 21.4 \\
GCN-LSTM \cite{yao2018exploring} & 77.3 &   -  &   -  & 36.8 & 27.9 & 57.0 & 116.3 & 20.9 & 80.5 &  -   &  -   & 38.2 & 28.5 & 58.3 & 127.6 & 22.0 \\
LBPF \cite{qin2019look}           & 77.8 &   -  &   -  & 37.4 & 28.1 & 57.5 & 116.4 & 21.2 & 80.5 &  -   &  -   & 38.3 & 28.5 & 58.4 & 127.6 & 22.0 \\
SGAE \cite{Yang:CVPR19}          & 77.6 &   -  &   -  & 36.9 & 27.7 & 57.2 & 116.7 & 20.9 & 80.8 &  -   &  -   & 38.4 & 28.4 & 58.6 & 127.8 & 22.1 \\
AoANet \cite{huang2019attentio}  & 77.4 &   -  &   -  & 37.2 & 28.4 & 57.5 & 119.8 & 21.3 & 80.2 &  -   &  -   & 38.9 & 29.2 & 58.8 & 129.8 & 22.4 \\
X-LAN \cite{xlinear} & 78.0 & 62.3 & 48.9 & 38.2 & 28.8 & 58.0 & 122.0 & 21.9 & 80.8 & 65.6 & 51.4 & 39.5 & 29.5 & 59.2 & 132.0 & 23.4 \\
Transformer \cite{sharma2018conceptual}   & 76.1 & 59.9 & 45.2 & 34.0 & 27.6 & 56.2 & 113.3 & 21.0 & 80.2 & 64.8 & 50.5 & 38.6 & 28.8 & 58.5 & 128.3 & 22.6 \\
X-Transformer \cite{xlinear} & 77.3 & 61.5 & 47.8 & 37.0 & 28.7 & 57.5 & 120.0 & 21.8 & 80.9 & 65.8 & 51.5 & 39.7 & 29.5 & 59.1 & 132.8 & 23.4 \\\hline

AutoCaption & 79.4 & 64.1 & 50.4 & 39.2 & 29.0 & 58.6 & 125.2 & 22.4 & 81.5 & 66.5 & 52.2 & 40.2 & 29.9 & 59.5 & 135.8 & 23.8  \\

AutoCaption (Vinvl) & \textbf{79.9} & \textbf{64.8} & \textbf{51.1} & \textbf{40.0} & \textbf{29.6} & \textbf{59.2} & \textbf{128.9} & \textbf{22.9} &
 
\textbf{82.3} & \textbf{67.6} & \textbf{53.3} & \textbf{41.2} & \textbf{30.4} & \textbf{60.4} & \textbf{139.5} & \textbf{24.3} 

\\\hline

			& \multicolumn{16}{c}{\textbf{Ensemble/Fusion}} \\ \hline
SCST \cite{rennie2017self}$^{\sum}$ & -   &   -  &   -  & 32.8 & 26.7 & 55.1 & 106.5 &   -  &  -   &  -   &  -   & 35.4 & 27.1 & 56.6 & 117.5 &  -   \\
RFNet \cite{jiang2018recurrent}$^{\sum}$ & 77.4 & 61.6 & 47.9 & 37.0 & 27.9 & 57.3 & 116.3 & 20.8 & 80.4 & 64.7 & 50.0 & 37.9 & 28.3 & 58.3 & 125.7 & 21.7 \\
GCN-LSTM \cite{yao2018exploring}$^{\sum}$& 77.4 & - & - & 37.1 & 28.1 & 57.2 & 117.1 & 21.1 & 80.9 &  -   &  -   & 38.3 & 28.6 & 58.5 & 128.7 & 22.1 \\
SGAE \cite{Yang:CVPR19}$^{\sum}$   &  -   &   -  &   -  &  -   &   -  &   -  &   -   &   -  & 81.0 &  -   &  -   & 39.0 & 28.4 & 58.9 & 129.1 & 22.2 \\
HIP \cite{yao2019hierarchy}$^{\sum}$ & -   &   -  &   - &  {38.0}  &  {28.6}  &  {57.8}  &  {120.3} & {21.4} & -   &   -  &   -  &  {39.1}  &  {28.9}  &  {59.2}  & {130.6}   & {22.3} \\
AoANet \cite{huang2019attentio}$^{\sum}$ & 78.7 & - & - & 38.1 & 28.5 & 58.2 & 122.7 & 21.7 & 81.6 &  -   &  -   & 40.2 & 29.3 & 59.4 & 132.0 & 22.8 \\
X-LAN$^{\sum}$ \cite{xlinear} & 78.8 & 63.4 & 49.9 & 39.1 & 29.1 & 58.5 & 124.5 & 22.2 & 81.6 & 66.6 & 52.3 & 40.3 & 29.8 & 59.6 & 133.7 & 23.6 \\
X-Transformer$^{\sum}$ \cite{xlinear} & 77.8 & 62.1 & 48.6 & 37.7 & 29.0 & 58.0 & 122.1 & 21.9 & 81.7 & 66.8 & 52.6 & 40.7 & 29.9 & 59.7 & 135.3 & 23.8 \\\hline

AutoCaption$^{\sum}$ & 79.8 & 64.7 & 51.2 & 40.3 & 29.6 & 59.2 & 128.5 & 22.8 & 
82.9 & 68.2 & 54.1 & 42.1 & 30.4 & 60.4 & 139.5 & 24.3 \\

AutoCaption$^{\sum}$(Vinvl) & \textbf{80.7} & \textbf{65.8} & \textbf{52.3} & \textbf{41.2} & \textbf{30.0} & \textbf{60.0} & \textbf{131.0} & \textbf{23.1} & 

\textbf{83.3} & \textbf{68.7} & \textbf{54.3} & \textbf{42.1} & \textbf{30.7} & \textbf{61.0} & \textbf{141.9} & \textbf{24.5} \\
		\Xhline{2\arrayrulewidth}
    \end{tabular}
	\vspace{-0.2in}
    \label{table:COCO}
\end{table*}

We sample as follows.
The controller first inputs the initial node No. 0 to predict the first node, the LSTM model outputs the hidden state, and then maps the probability distribution to the entire search space through the fully connected layer. 
Then the controller samples an activation function.
In Figure \ref{fig:autornn}, the controller chooses the $Tanh$ activation function, which means that node $1$ of the recurrent cell should compute 
\begin{equation}
h_1 = Tanh{(x_t \cdot W^{(x)} + h_{t-1} \cdot W^{(h)}_1)}.
\end{equation}
The controller uses the $Tanh$ layer as an input to predict the connection between the next node and the current node. 
In Figure \ref{fig:autornn}, the controller predicts  node $1$ at the Node $2$ stage, which represents the connection between node $2$ and node $1$, and then uses node $1$ as an input to predict the choice of the node $2$ layer.
The controller generate the activation function $\text{Elu}$. 
Thus, node $2$ of the cell computes 
\begin{equation}
h_2 = \text{Elu}(h_1 \cdot W^{(h)}_{2, 1}).
\end{equation}
The controller again samples a connection and an activation function.
In our example, it chooses connection $2$ and the activation function $\text{ReLU}$. Therefore, 
\begin{equation}
h_3 = \text{ReLU}(h_2 \cdot W^{(h)}_{3, 2}).
\end{equation}
The whole process loops and iterates until all connections and nodes are determined. 
Finally, a recurrent neural network is generated.
We use average all the output of the leaf nodes as our final output, and the leaf nodes that are not selected as inputs to any other nodes.

When a recurrent neural network is generated, it needs to be trained on the training set to evaluate the performance of the selected network. 
In order to evaluate the selected network more quickly, less epoch training can be performed. 
After the training is completed, it needs to be evaluated by means of reinforcement learning strategy gradient. 
The pre-trained sub-network is evaluated on the verification set to get the reward of the currently selected model. 
For models that are superior to the benchmark reward, reinforcement learning performs positive optimization on the controller and negatively optimizes the model that is lower than the benchmark reward, which ultimately makes the controller's ability to generate models continuously improve.

In order to speed up the search speed of the entire network, we use a strategy of sharing parameters. 
First, a super network is defined, which contains all possible connection methods of all sub-networks and shares parameters with the sub-networks.
The parameters of the generated sub-networks are directly inherited from the super-network. 
At the training phrase of the sub-networks, the parameters of the super-network are also updated synchronously while the parameters of the sub-networks are updated. This can effectively avoid repeated initialization of the sub-networks and accelerate the search speed of the network.

The network structure search method of reinforcement learning can ensure the diversity of the generated network, which is conducive to selecting the network with more excellent performance.
The entire network architecture search is performed on the image captioning  dataset MSCOCO and
it uses the image features extracted in advance, so that only the text generation model needs to be searched for the network structure, which is beneficial to speed up the network search speed.

\section{Experiments}

\subsection{Dataset and Implementation Details}

We use the MSCOCO dataset\cite{lin2014microsoft} which is now the most used benchmark dataset of the image captioning task to evaluate our model. 
The official dataset of the MSCOCO includes 82,783 training images, 40,504 validation images and 40,775 test images.
We use both the offline evaluation and the online evaluation to verify our proposed model.
For offline evaluation, we use the data split as in \cite{Karpathy2015Deep} which is usually used as offline evaluation in recent papers. 
The offline dataset train set contains 113,287 images, each with 5 captions.
We use a set of 5000 images for validation and report the results on Karpathy splits \cite{vinyals2017show} with 5000 images.

For preprocessing of captions, we transform all letters in the captions to lowercase and remove all the non-alphabetic characters. 
We only use the word which count more than 5 and we get a vocabulary which has 9487 words in the end. 
Words occur less than five times are replaced with an unknown token $<UNK>$. 
We statistics the length of all captions and find that 97.7\% of caption length is less than 16 and more longer will make the training of the LSTM become more difficult, so we truncate all the captions longer than 16 tokens.
For caption text data, We turn each word into ``one-hot" vector.

We leverage the off-the-shelf Faster-RCNN pre-trained on ImageNet \cite{ImageNet} and Visual Genome \cite{krishna2017visual} to extract image region features \cite{anderson2017bottom}. 
The whole image captioning architecture are mainly implemented with PyTorch, optimized with Adam \cite{kingma2014adam}.

For the training stage, we follow the training schedule in \cite{vaswani2017attention} to optimize the whole architecture with cross-entropy loss. 
The warmup steps are set as 10,000 and the mini-batch size is 50. 
We set the maximum number of iterations as 100 to select the best training model.
At the training phrase, the model is trained with self-critical training strategy, as in \cite{rennie2017self, zhu2018as}.
We first choose the initial model, which is trained with cross-entropy loss and gets the best cider score on the validation set.
Then, we set the learning rate to 0.00001 and the maximum number of iterations as 100 epochs, we use CIDEr-reward to further optimize the entire architecture.
At the inference stage, we adopt the beam search strategy and set the beam size as 3.

In order to verify the performance of our proposed model further, We further combined our AutoRNN with bilinear attention \cite{xlinear}, and we called the model as the AutoCaption.

We evaluate our model using five metrics which are widely used in image captioning task, including \textbf{CIDEr} \cite{Vedantam2015CIDEr}, \textbf{BLEU} \cite{Papineni2002BLEU}, \textbf{METEOR} \cite{Denkowski2014Meteor}, \textbf{ROUGE-L} \cite{lin2004rouge} and \textbf{SPICE} \cite{spice2016}.

\begin{table*}[!tb]\scriptsize
  \centering
  \caption{Leaderboard of the published state-of-the-art image captioning models on the MSCOCO online testing server, where B@$N$, M, R, and C are short for BLEU@$N$, METEOR, ROUGE-L, and CIDEr scores. All values are reported as percentage (\%).}
  \label{table:leaderboard}
  \vspace{-0.00in}
  \begin{tabular}{l|*{13}{c|}c}
  \Xhline{2\arrayrulewidth}
      \multicolumn{1}{c|}{\multirow{2}{*}{{Model}}} & \multicolumn{2}{c|}{{B@1}} & \multicolumn{2}{c|}{{B@2}} & \multicolumn{2}{c|}{{B@3}} & \multicolumn{2}{c|}{{B@4}} & \multicolumn{2}{c|}{{M}} & \multicolumn{2}{c|}{{R}} & \multicolumn{2}{c}{{C}} \\\cline{2-15}
      \multicolumn{1}{c|}{}&c5 &c40 &c5 &c40 &c5 &c40&c5 &c40&c5 &c40&c5 &c40&c5 &c40 \\\hline
      \hline
       {LSTM-A} (ResNet-152) \cite{yao2017boosting}     & 78.7 & 93.7 & 62.7 & 86.7 & 47.6 & 76.5 & 35.6 & 65.2 & 27.0 & 35.4 & 56.4 & 70.5 & 116.0 & 118.0  \\\hline
	    {Up-Down} (ResNet-101) \cite{anderson2017bottom} & 80.2 & 95.2 & 64.1 & 88.8 & 49.1 & 79.4 & 36.9 & 68.5 & 27.6 & 36.7 & 57.1 & 72.4 & 117.9 & 120.5  \\\hline
      {RFNet} (ResNet+DenseNet+Inception) \cite{jiang2018recurrent}  & 80.4 & 95.0 & 64.9 & 89.3 & 50.1 & 80.1 & 38.0 & 69.2 & 28.2 & 37.2 & 58.2 & 73.1 & 122.9 & 125.1 \\\hline
			{SGAE} (ResNet-101) \cite{Yang:CVPR19}           & 81.0 & 95.3 & 65.6 & 89.5 & 50.7 & 80.4 & 38.5 & 69.7 & 28.2 & 37.2 & 58.6 & 73.6 & 123.8 & 126.5 \\\hline
			{GCN-LSTM} (ResNet-101) \cite{yao2018exploring}  & 80.8 & 95.2 & 65.5 & 89.3 & 50.8 & 80.3 & 38.7 & 69.7 & 28.5 & 37.6 & 58.5 & 73.4 & 125.3 & 126.5 \\\hline
			{AoANet} (ResNet-101) \cite{huang2019attentio}   & 81.0 & 95.0 & 65.8 & 89.6 & 51.4 & 81.3 & 39.4 & 71.2 & 29.1 & 38.5 & 58.9 & 74.5 & 126.9 & 129.6 \\\hline
			{HIP (SENet-154) \cite{yao2019hierarchy}}                    & 81.6 & 95.9 & 66.2 & 90.4 & 51.5 & 81.6 & 39.3 & 71.0 & 28.8 & 38.1 & 59.0 & 74.1 & 127.9 & 130.2 \\\hline
			{X-LAN (ResNet-101) \cite{xlinear}}                & 81.1 & 95.3 & 66.0 & 89.8 & 51.5 & 81.5 & 39.5 & 71.4 & 29.4 & 38.9 & 59.2 & 74.7 & 128.0 & 130.3 \\\hline
            {X-LAN (SENet-154) \cite{xlinear}}                 & 81.4 & 95.7 & 66.5 & 90.5 & 52.0 & 82.4 & 40.0 & 72.4 & 29.7 & 39.3 & 59.5 & 75.2 & 130.2 & 132.8 \\\hline
			{X-Transformer (ResNet-101) \cite{xlinear}}                & 81.3 & 95.4 & 66.3 & 90.0 & 51.9 & 81.7 & 39.9 & 71.8 & 29.5 & 39.0 & 59.3 & 74.9 & 129.3 & 131.4 \\\hline
			{X-Transformer (SENet-154) \cite{xlinear}}                 & 81.9 & 95.7 & 66.9 & 90.5 & 52.4 & 82.5 & 40.3 & 72.4 & 29.6 & 39.2 & 59.5 & 75.0 & 131.1 & 133.5 \\\hline\hline
			{AutoCaption (ResNeXt-152)}                 & 82.5 & 96.2 & 67.7 & 91.4 & 53.2 & 83.6 & 41.1 & 73.9 & 30.0 & 39.8 & 60.1 & 75.8 & 133.3 & 135.7 \\\hline
			{AutoCaption (Vinvl)}                 & \textbf{82.5} & \textbf{96.6} & \textbf{67.8} & \textbf{91.9} & \textbf{53.3} & \textbf{84.2} & \textbf{41.1} & \textbf{74.3} & \textbf{30.3} & \textbf{40.1} & \textbf{60.4} & \textbf{76.0} & \textbf{135.9} & \textbf{138.9} \\
			
      \Xhline{2\arrayrulewidth}
  \end{tabular}
  \vspace{-0.15in}
\end{table*}

\subsection{Evaluation}

To better quantify the effect of our AutoCaption method, we evaluate our full model against prior X-Linear posted in \cite{xlinear}. 
We regard the X-Linear model as our baseline model. 
The CNN used here is the ResNeXt \cite{Xie2016Aggregated} which is showing better results than ResNet \cite{He2016Identity}. We further combined the recently proposed Vinvl \cite{zhang2021vinvl} method to further verify the performance of the model. 

\noindent\textbf{Offline Evaluation.}
In Table \ref{table:COCO}, we report the performance of our AutoCaption model without reinforce learning, the AutoCaption model with reinforce learning on test portion of the Karpathy splits.
Through the evaluation metrics we can find that our model can improve all the performance and get better performance. 
This results further verifies the effectiveness of the AutoCaption model automatically designed by NAS.
The CIDEr score of our single-model AutoCaption with Vinvl feature can achieve 139.5\% with CIDER score optimization. 
This score is higher than a variety of hand-designed image captioning methods. 
Furthermore, we use model ensemble technology to ensemble multiple models of different architectures to evaluate our AutoCaption model further. Our method can achieve a higher score, reaching 141.9 on the CIDEr score.

Through the experimental results, we can find that our AutoCaption model is better than a variety of baseline methods, further verifying the effectiveness of the automatically designed model.

\noindent\textbf{Online Evaluation.}
Then we use our best model to get the test result with the official test split, and submit our result to the official MSCOCO evaluation server. 
In addition, we evaluate our AutoCaption on the official testing set by submitting the ensemble versions to online testing server.
Table \ref{table:leaderboard} details the performances over official testing images with 5 reference captions (c5) and 40 reference captions (c40).
Note that here we adopt backbone ResNeXt-152 and Vinvl for online evaluation. 
The results show that our AutoCaption can achieve better performance than other state-of-the-art methods.

\subsection{Quantitative Analysis}

\begin{figure*}[!tb]
\center
\includegraphics[width=6.5in]{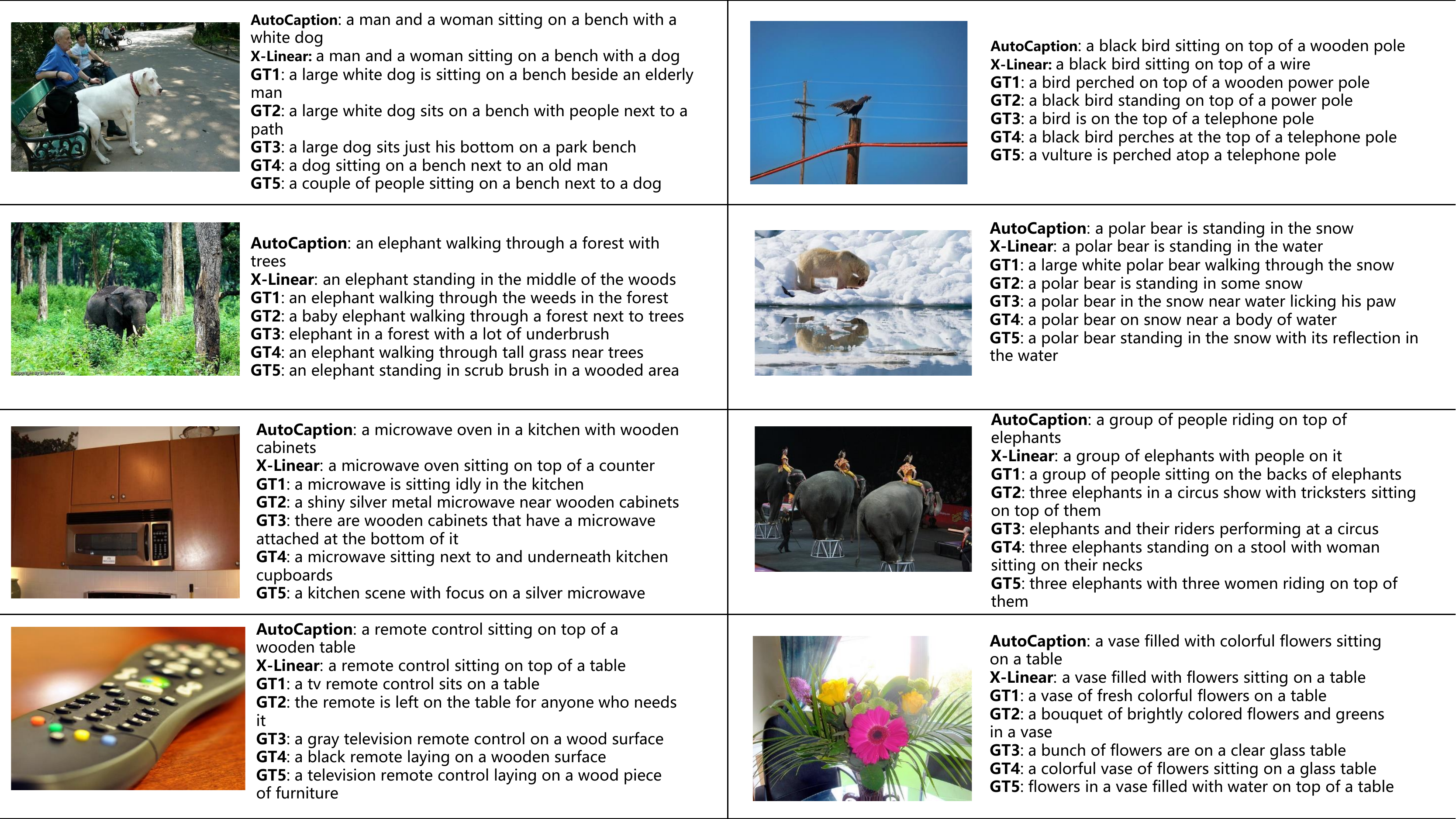}
\caption{Examples of image captioning results by AutoCaption and X-Linear \cite{xlinear}, coupled with the corresponding ground truth sentences.}
\label{fig:results}
\end{figure*}

Figure \ref{fig:results} showcases several image captioning results of AutoCaption, coupled with human-annotated ground truth sentences (GT).
Compared with the captions of baseline model X-Linear, our Auto-Caption is more accurate and contains richer attribute information.
For example, AutoCaption generates the sentence “a man and a woman sitting on a bench with a white dog” in the first image, which can describe the dog with the attribute “white”, but X-Linear can not generate the attribute word.
This verifies that our method can more accurately describe the attribute information of the object in the image.
For another example, AutoCaption can describe the image with the sentence “a black bird sitting on top of a wooden pole”, but the X-Linear recognize the object error, and describe the image with the wrong word “wire”.
This again confirms the advantage of our AutoCaption model generating more accurate description sentences.
Other examples also show similar results, which further verified that the automatically designed model can achieve better results with the manually designed model.

\subsection{Experimental Analysis}

\begin{table}[!tb]\scriptsize
\centering
  \caption{\small Study on the different search blocks of the AutoCaption, where B@$N$, R, and C are short for BLEU@$N$, ROUGE-L, and CIDEr. All values are reported as percentage (\%).}
\label{table:block}
\begin{tabular}{l|l|cccccccc}
\Xhline{2\arrayrulewidth}
Model & Blocks & B@1 & B@2 & B@3 & B@4 & R & C \\ \hline\hline
ShowTell \cite{vinyals2017show} & - & 70.3 & 53.2 & 38.9 & 28.2 & 52.8 & 93.4 \\ \hline
AutoCaption & 10 & 70.0 & 53.1 & 39.1 & 28.6 & 53.0 & 93.8 \\ \hline
AutoCaption & 8  & 70.4 & 53.4 & 39.2 & 28.7 & 53.3 & 95.0 \\ \hline
AutoCaption & 6  & \textbf{70.6} & \textbf{53.9} & \textbf{39.9} & \textbf{29.3} & \textbf{53.4} & \textbf{95.9} \\ \hline
\Xhline{2\arrayrulewidth}
\end{tabular}
\end{table}

\begin{table}[!tb]\scriptsize
\centering
  \caption{\small Study on the whether to use metric evaluation method in the AutoCaption train phrase, where B@$N$, R, and C are short for BLEU@$N$, ROUGE-L, and CIDEr. All values are reported as percentage (\%).}
\label{table:eval}
\begin{tabular}{l|l|cccccccc}
\Xhline{2\arrayrulewidth}
Model & Metric Eval & B@1 & B@2 & B@3 & B@4 & R & C \\ \hline\hline
AutoCaption & False & 70.2 & 53.2 & 39.1 & 28.5 & 53.0 & 94.4 \\ \hline
AutoCaption & True& 70.6 & 53.9 & 39.9 & 29.3 & 53.4 & 95.9	 \\ \hline
\Xhline{2\arrayrulewidth}
\end{tabular}
\end{table}

In Table \ref{table:block}, we further discuss the influence of the number of different blocks on the model performance.
All models int Table \ref{table:block} use the bottom-up features \cite{anderson2017bottom}, and trained with only the cross-entropy loss, and not trained with the CIDEr reward. 
None of these methods use beam search method for evaluation.
We can find that our AutoCaption models achieve better performance than LSTM method
and when the blocks is 6, the model achieves the best performance.
A smaller block achieves better performance. 
This is most likely due to the limitation of MSCOCO dataset.
A larger block may have a certain degree of overfitting without achieving better performance.
This verifies the effectiveness of our proposed method, which can correctly and automatically design an effective model and achieve better performance.

\begin{table}[!tb]\scriptsize
\centering
  \caption{\small Model size comparison of LSTM, Transformer, AutoRNN-6, AutoRNN-8 and AutoRNN-10 methods.}
\label{table:size}
\begin{tabular}{l|l|ccc}
\Xhline{2\arrayrulewidth}
Model & Hidden Size & Model Size & Params \\ \hline\hline
LSTM & 512 & 8.0M & 2.0M \\ \hline
Transformer & 512 & 168.4M & 42.1M \\ \hline
AutoRNN-6 & 512 & 14.0M & 3.5M \\ \hline
AutoRNN-8 & 512 & 18.0M & 4.5M \\ \hline
AutoRNN-10 & 512 & 22.0M & 5.5M \\ \hline\hline
LSTM & 1024 & 32.0M & 8.0M \\ \hline
Transformer & 1024 & 480.7M & 120.2M \\ \hline
AutoRNN-6 & 1024 & 56.0M & 14.0M \\ \hline
AutoRNN-8 & 1024 & 72.0M & 18.0M \\ \hline
AutoRNN-10 & 1024 & 88.0M & 22.0M \\ \hline
\Xhline{2\arrayrulewidth}
\end{tabular}
\end{table}

In Table \ref{table:size}, we compare the sizes of different models.
We can find that the Transformer model has larger size than the traditional the LSTM model. 
In recent works, the image captioning model with the Transformer \cite{xlinear} has achieved the better performance.
However, this will lead to larger model size and has impact deployment to embedded devices.
The AutoRNN-$N$ represents the model has $N$ number of blocks.
We can find that AutoRNN is only a little larger than the traditional LSTM model, but it performs better than LSTM and Transformer both in image capture tasks
as shown in Table \ref{table:COCO} and \ref{table:leaderboard}.
Through AutoCaption, we can flexibly control the size of the model to meet the requirements of different devices.

In Table \ref{table:eval}, we compare whether to use evaluation metric in training phrase to evaluate the model sampled by the controller.
The original method evaluates the model through loss, which does not effectively verify the performance of the model. 
We use the CIDEr metric of image captioning to directly evaluate models sampled by the controller during the training and test phase.
In the comparison, it can be found that the evaluation-based method can select sub-models with better performance.

\section{Conclusion}

We present a novel AutoCaption for image captioning, which use the Neural Architecture Search method to find better decoder to generate better content.
In order to improve the performance of text generation in image captioning model, a network structure search method for text generation based on reinforcement learning is designed.
This method can accelerate the design speed of the text generation model, and allows the text generation network to take into account the characteristics of the text in the image captioning task.
Extensive experiments conducted on MSCOCO dataset demonstrate the performance of the AutoCaption method.
The results of the paper also verify that the automatically designed network can achieve better performance on the multi-modal task Image Captioning than the manually designed network.

{\small
\bibliographystyle{ieee_fullname}
\bibliography{egbib}
}

\end{document}